\pdfoutput=1
\documentclass[10pt,twocolumn,letterpaper]{article}

\usepackage{cvpr}
\usepackage{times}
\usepackage{epsfig}
\usepackage{graphicx}
\usepackage{amsmath}
\usepackage{amssymb}
\usepackage{multirow}
\usepackage{booktabs} 
\usepackage{algorithm}
\usepackage{algorithmic}
\usepackage{subfigure}

\usepackage[pagebackref=true,breaklinks=true,letterpaper=true,colorlinks,bookmarks=false]{hyperref}

\cvprfinalcopy 

\def\cvprPaperID{279} 

\usepackage{xcolor} 

\ifcvprfinal\fi
\begin{document}

\title{Deep Variation-structured Reinforcement Learning for Visual Relationship and Attribute Detection}

\author{Xiaodan~Liang \quad Lisa Lee \quad Eric P. Xing \\
School of Computer Science, Carnegie Mellon University\\
\{xiaodan1,lslee,epxing\}@cs.cmu.edu
}

\maketitle

\begin{abstract}
	
Despite progress in visual perception tasks such as image classification and detection, computers still struggle to understand the interdependency of objects in the  scene as a whole, e.g., relations between objects or their attributes. Existing methods often ignore global context cues capturing the interactions among different object instances, and can only recognize a handful of types by exhaustively training individual detectors for all possible relationships. To capture such global interdependency, we propose a deep Variation-structured Reinforcement Learning (VRL) framework to sequentially discover object relationships and attributes in the whole image. First, a directed semantic action graph is built using language priors to provide a rich and compact representation of semantic correlations between object categories, predicates, and attributes. Next, we use a variation-structured traversal over the action graph to construct a small, adaptive action set for each step based on the current state and historical actions. In particular, an ambiguity-aware object mining scheme is used to resolve semantic ambiguity among object categories that the object detector fails to distinguish. We then make sequential predictions using a deep RL framework, incorporating global context cues and semantic embeddings of previously extracted phrases in the state vector. Our experiments on the Visual Relationship Detection (VRD) dataset and the large-scale Visual Genome dataset validate the superiority of VRL, which can achieve significantly better detection results on datasets involving thousands of relationship and attribute types. We also demonstrate that VRL is able to predict unseen types embedded in our action graph by learning correlations on shared graph nodes.
\end{abstract}

\section{Introduction}

\begin{figure}[!tp]
	\begin{center}
		\includegraphics[scale=0.33]{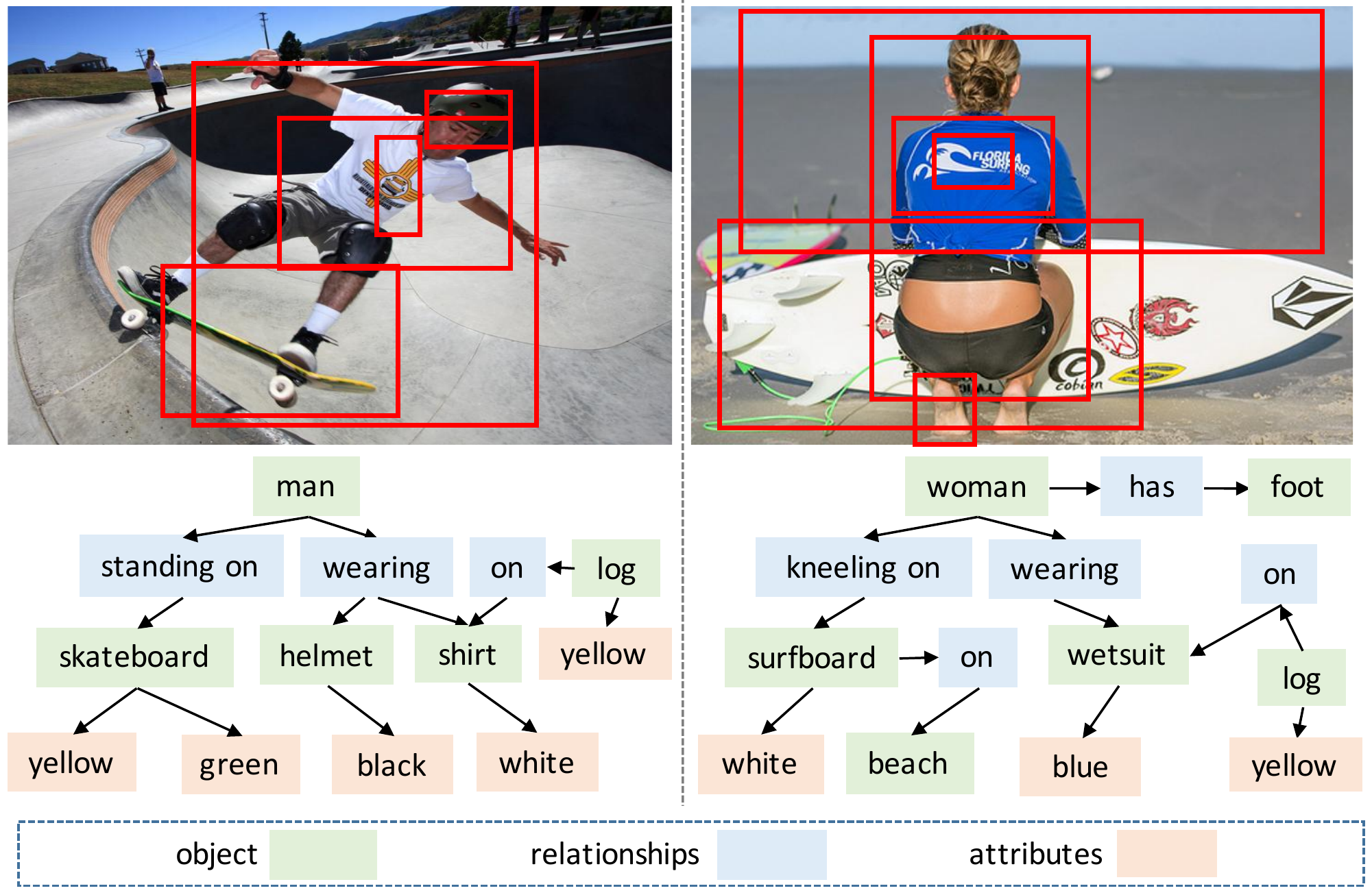}
		\caption{{In each example (left and right), we show the bounding boxes of objects in the image (top), and the relationships and attributes recognized by our proposed VRL framework (bottom). Only the top few results are illustrated for clarity.}}
		\label{fig:task}
	\end{center}
	\vspace{-8mm}
\end{figure}

Although much progress has been made in image classification~\cite{he2015deep}, detection~\cite{ren2015faster} and segmentation~\cite{long2015fully}, we are still far from reaching the goal of holistic scene understanding---that is, a model capable of recognizing the interactions and relationships between objects, and describing their attributes. While objects are the core building blocks of an image, it is often the relationships and attributes that determine the holistic interpretation of the scene. For example in Fig.~\ref{fig:task}, the left image can be understood as ``a man standing on a yellow and green skateboard", {and} the right image as ``a woman wearing a blue wet suit and kneeling on a surfboard". Being able to extract
and exploit such visual information would benefit many
real-world applications such as image search~\cite{qin2013query}, question answering~\cite{antol2015vqa,johnson2015densecap}, and fine-grained recognition~\cite{zhu2014reasoning, farhadi2009describing}.

\begin{figure*}[!tp]
	\begin{center}
		\includegraphics[scale=0.51]{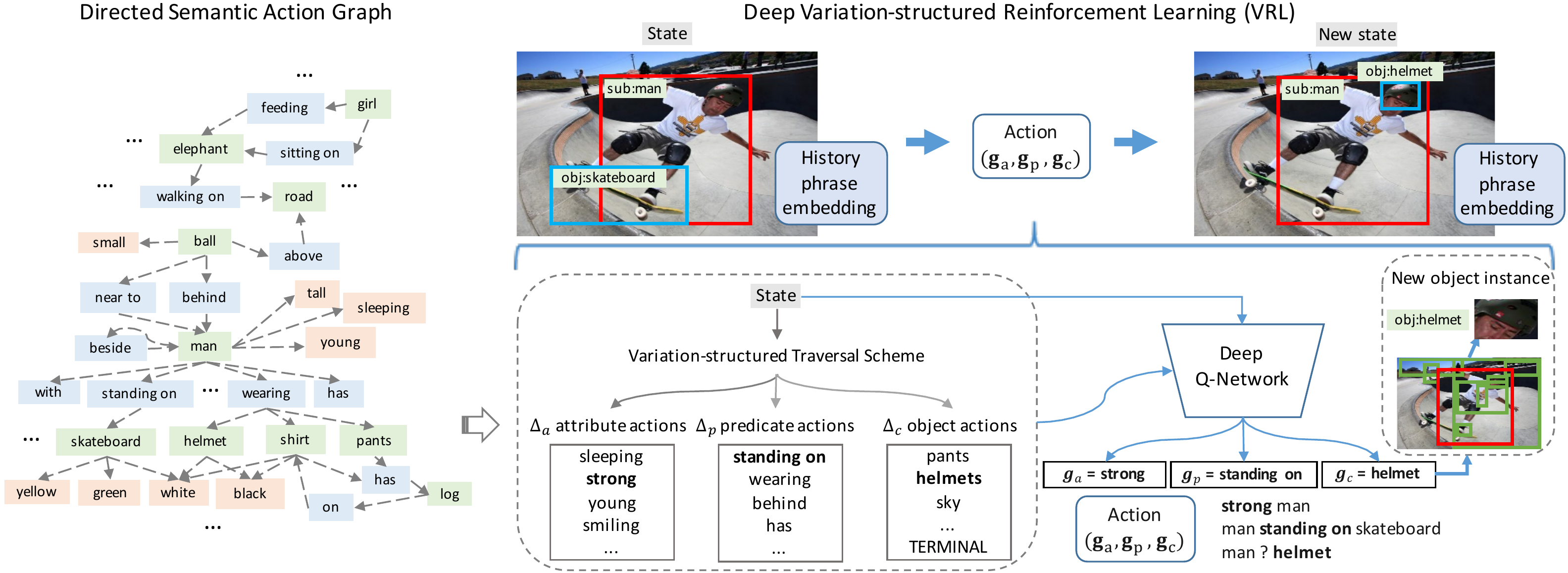}
		\caption{{An overview of {the} VRL framework that sequentially detects relationships (``subject-predicate-object") and attributes (``subject-attribute"). {First, we} build a directed semantic action graph $\mathcal{G}$ to configure the whole action space. In each step, the {input state consists of the current subject and object instances (``sub:man'', ``obj:skateboard'') and a history phrase embedding, which captures the search paths that have already been traversed by the agent. A variation-structured traversal scheme over $\mathcal{G}$ dynamically constructs three small action sets $\Delta_a$, $\Delta_p$, $\Delta_c$. The agent predicts three actions: (1) $\mathbf{g}_a \in \Delta_a$, an attribute of the subject; (2) $\mathbf{g}_p \in \Delta_p$, the predicate between the subject and object; and (3) $\mathbf{g}_c \in \Delta_c$, the next object category of interest (``obj:helmet''). The new state consists of the new subject/object instances (``sub:man'', ``obj:helmet'') and an updated history phrase embedding.}}}
		\label{fig:framework}
	\end{center}
	\vspace{-8mm}
\end{figure*}

Visual relationships are a pair of localized objects connected via a predicate; for example, predicates can be actions (``kick''), comparative (``smaller than''), spatial (``near to''), verbs (``wear''), or prepositions (``with''). Attributes describe a localized object, e.g., with color (``yellow'') or state (``standing''). Detecting relationships and attributes is more challenging than traditional object detection~\cite{ren2015faster} due to the following reasons: (1) There are a massive {number of} possible relationship and attribute types (e.g., 13,894 relationship types in Visual Genome~\cite{krishna2016visual}), resulting in a greater skew of rare and infrequent types. (2) Each object can be associated with many relationships and attributes, making it inefficient to exhaustively search all { possible relationships} for each pair of objects. (3) A global, {holistic} perspective of the image is essential to resolve semantic {ambiguities} (e.g., ``woman wearing wetsuit'' vs. ``woman wearing shirt''). Existing approaches~\cite{sadeghi2011recognition,johnson2015image, lu2016visual} only predict a limited set of {relationship} types (e.g., 13 in Visual Phrase~\cite{sadeghi2011recognition}) and ignore semantic interdependencies between relationships and attributes by evaluating each region {within a scene separately}~\cite{lu2016visual}. It is impractical to exhaustively search all possibilities for each region, and also deviates from human perception. Therefore, it is preferable to have a more principled decision-making framework, which can discover all relevant relationships and attributes within a small number of search steps. To address the aforementioned issues, we propose a deep Variation-structured Reinforcement Learning (VRL) framework which sequentially detects relationship and attribute instances by exploiting global context cues.

First, we use language priors to build a directed semantic action graph $\mathcal{G}$, where the nodes are nouns, attributes, and predicates, connected by directed edges that represent semantic correlations (see Fig.~\ref{fig:framework}). This graph provides a highly-informative, compact representation that enables the model to learn rare relationships and attributes  from frequent ones using shared graph nodes. For example, the semantic meaning of ``riding'' learned from ``person-riding-bicycle'' can help predict the rare phrase ``child-riding-elephant''. This generalizing ability allows VRL to handle a considerable number of possible {relationship} types.

Second, existing deep reinforcement learning (RL) models~\cite{silver2016mastering} often require several costly episodes of trial and error to converge, even with a small action space, and our large action space would exacerbate this problem. To efficiently discover all relationships and attributes in a small number of steps, we introduce a novel variation-structured traversal scheme over the action graph which constructs small, adaptive action sets $\Delta_a, \Delta_p, \Delta_c$ for each step based on the current state and historical actions: $\Delta_a$ contains candidate attributes to describe an object; $\Delta_p$ contains candidate predicates for relating a pair of objects; and $\Delta_c$ contains new object instances to mine in the next step. Since an object instance may belong to multiple object categories which the object detector cannot distinguish, we introduce an ambiguity-aware object mining scheme to assign each object with the most appropriate category given the global scene context.  Our variation-structured traversal scheme offers a very promising technique for extending the applications of deep RL to complex real-world tasks.

Third, to incorporate global context cues for better reasoning, we explicitly encode the semantic embeddings of previously extracted phrases in the state vector. It {makes a better} tradeoff between increasing the input dimension and utilizing more historical context, compared to appending history frames~\cite{zhu2016target} or binary action vectors~\cite{caicedo2015active} as in previous RL methods.  

Extensive experiments on the Visual Relationship Detection (VRD) dataset~\cite{lu2016visual} and Visual Genome dataset~\cite{krishna2016visual} demonstrate that the proposed VRL outperforms {state-of-the-art methods} for both relationship and attribute detection, and also has good generalization capabilities for predicting unseen types.

\section{Related Works}

\textbf{Visual relationship and attribute detection.} {There has been an increased interest in the problem of visual relationship detection}~\cite{sadeghi2011recognition,sadeghi2015viske,krishna2016visual}. However, most existing approaches~\cite{sadeghi2011recognition}~\cite{krishna2016visual} can detect only a handful of pre-defined, frequent types by training individual detectors for each relationship. Recently, Lu et al.~\cite{lu2016visual} leveraged word embeddings to handle large-scale relationships. However, their model still ignores the structured correlations between objects and relationships. Furthermore, some methods~\cite{johnson2015image,schuster2015generating,liao2016support} organized predictions into a scene graph which can provide a structured representation for describing the objects, their attributes and relationships in each image. In particular, Johnson et al.~\cite{johnson2015image} introduced a conditional random field model for reasoning about possible groundings of scene graphs while Schuster et al.~\cite{schuster2015generating} proposed a rule-based and classifier-based scene graph parser. In contrast, the proposed VRL makes the first attempt to sequentially discover objects, relationships and attributes by fully exploiting global interdependency. 

\textbf{Deep reinforcement learning.} Integrating deep learning methods with reinforcement learning {(RL)} ~\cite{kaelbling1996reinforcement} has recently shown very promising results on decision-making problems. For example, Mnih et al.~\cite{mnih2015human} proposed using deep Q-networks to play ATARI games. Silver et al.~\cite{silver2016mastering} proposed a new search algorithm based on the integration of Monte-Carlo tree search with deep RL, which {beat} the world champion in the game of Go. 
{Other efforts applied deep RL to various real-world tasks,} e.g., robotic manipulation~\cite{gu2016deep}, indoor navigation~\cite{zhu2016target}, and object proposal generation~\cite{caicedo2015active}. Our work deals with real-world scenes that are much more  complex than ATARI games or images taken in some constrained scenarios, and investigates how to make decisions over a larger action space (e.g., thousands of attribute types). To handle such a large action space, we propose a variation-structured traversal scheme over the whole action graph to decrease the number of possible actions in each step, which substantially reduces the number of trials and thus speeds up the convergence.

\section{Deep Variation-structured Reinforcement Learning}

We propose a novel VRL framework which formulates the problem of detecting visual relationships and attributes as a sequential decision-making process. An overview is provided in Fig.~\ref{fig:framework}. The key components of VRL, including the directed semantic action graph, the variation-structured traversal scheme, the state space, and the reward function, are detailed in the following sections. 


\begin{figure*}[!tp]
	\begin{center}
		\includegraphics[scale=0.58]{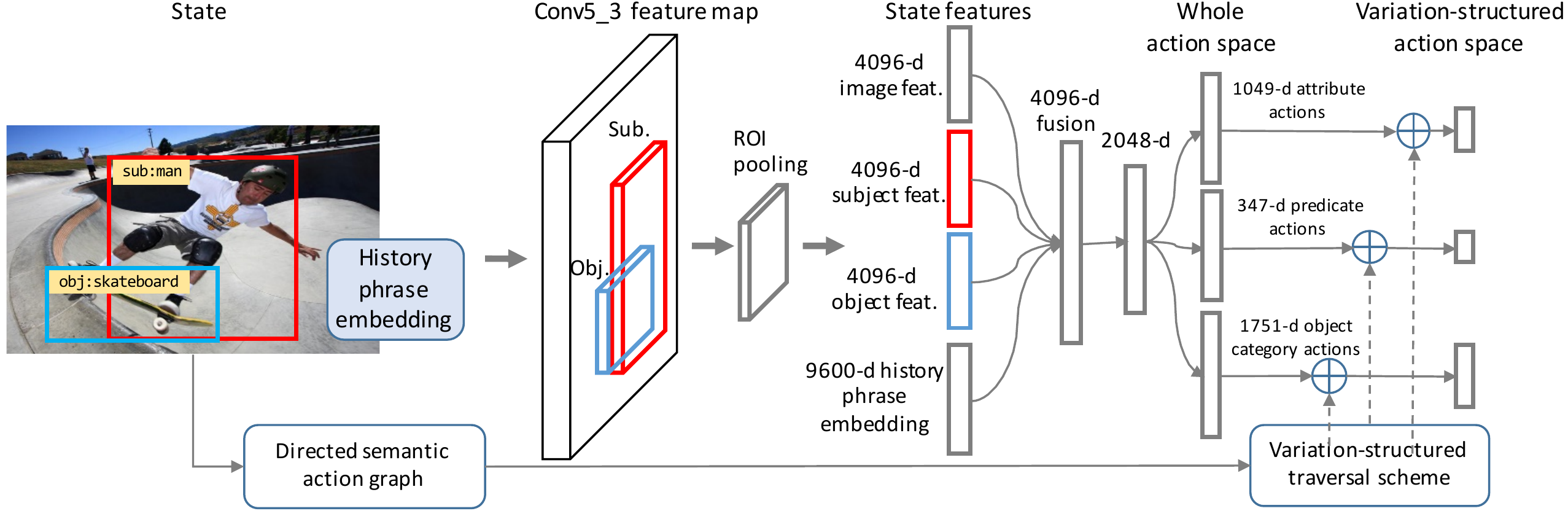}
		\caption{Network architecture of {deep VRL. The state vector $\bf{f}$ is a concatenation of (1) a 4096-dim feature of the whole image, taken from the fc6 layer of the pre-trained VGG-16 ImageNet model~\cite{simonyan2014very}; (2) two 4096-dim features of the subject $s$ and object $s'$ instances, taken from the conv5\_3 layer of the trained Faster R-CNN object detector; and (3) a 9600-dim history phrase embedding, which is created by concatenating four 2400-dim semantic embeddings from a Skip-thought language model~\cite{kiros2015skip} of the last two relationship phrases (relating $s$ and $s'$) and the last two attribute phrases (describing $s$) that were predicted by VRL. A variation-structured traversal scheme over the directed semantic action graph produces a smaller action space from the whole action space, which originally consists of $|\mathcal{A}| = 1049$ attributes, $|\mathcal{P}| = 347$ predicates, and $|\mathcal{C}| = 1750$ object categories plus one terminal trigger. From this variation-structured action space, the model selects actions with the highest predicted Q-values in state $\bf{f}$. }
		}
		\label{fig:network}
	\end{center}
	\vspace{-6mm}
\end{figure*}

\vspace{-2mm}
\subsection{Directed Semantic Action Graph}

We build a directed semantic graph $\mathcal{G} = (\mathcal{V}, \mathcal{E})$ to organize all possible object nouns, attributes, and relationships into a compact and semantically meaningful representation (see Fig.~\ref{fig:framework}). The nodes $\mathcal{V}$ consist of the set of all candidate object categories $\mathcal{C}$, attributes $\mathcal{A}$, and predicates $\mathcal{P}$. Object categories in $\mathcal{C}$ are nouns, and may be people, places, or parts of objects. Attributes in $\mathcal{A}$ can describe color, shape, or pose. Relationships are directional, i.e. they relate a subject noun and an object noun via a predicate. Predicates in $\mathcal{P}$ can be spatial (e.g., ``inside of''), compositional (e.g. ``part of'') or action (e.g., ``swinging''). 

The directed edges $\mathcal{E}$ consist of attribute phrases {$\mathcal{E}_\mathcal{A} \subseteq \mathcal{C} \times \mathcal{A}$ and predicate phrases $\mathcal{E}_\mathcal{P} \subseteq \mathcal{C} \times \mathcal{P} \times \mathcal{C}$}. An attribute phrase $(c, a) \in \mathcal{E}_\mathcal{A}$ represents an attribute $a \in A$ belonging to a noun $c \in \mathcal{C}$. For example, the attribute phrase ``young girl'' can be represented by (``girl'', ``young'') $\in \mathcal{E}_\mathcal{A}$. A predicate phrase $(c, p, c') \in \mathcal{E}_\mathcal{P}$ represents a subject noun $c \in \mathcal{C}$ and an object noun $c' \in \mathcal{C}$ related by a predicate $p \in P$. For example, the predicate phrase ``a man is swinging a bat'' can be represented by (``man'', ``swinging'', ``bat'') $\in \mathcal{E}_\mathcal{P}$.

The recently released Visual Genome dataset~\cite{krishna2016visual} provides a large-scale annotation of images containing 18,136 unique object categories, 13,041 unique attributes, and 13,894 unique relationships. We then select the types that appear at least 30 times in Visual Genome dataset, resulting in 1,750 object-, 8,561 attribute-, and 13,823 relationship-types. From these attribute and relationship types, we build a directed semantic action graph by extracting all unique object category words, attribute words, and predicate words as the graph nodes. Our directed action graph thus contains $|\mathcal{C}| = 1750$ object nodes, $|\mathcal{A}| = 1049$ attribute nodes, and $|\mathcal{P}| = 347$ predicate nodes. On average, each object word is connected to $5$ attribute words and $15$ predicate words. This semantic action graph serves as the action space for VRL, as we will see in the next section.\\

\vspace{-2mm}
\subsection{Variation-structured RL}

 \begin{figure}[!tp]
 	\begin{center}
 		\includegraphics[scale=0.34]{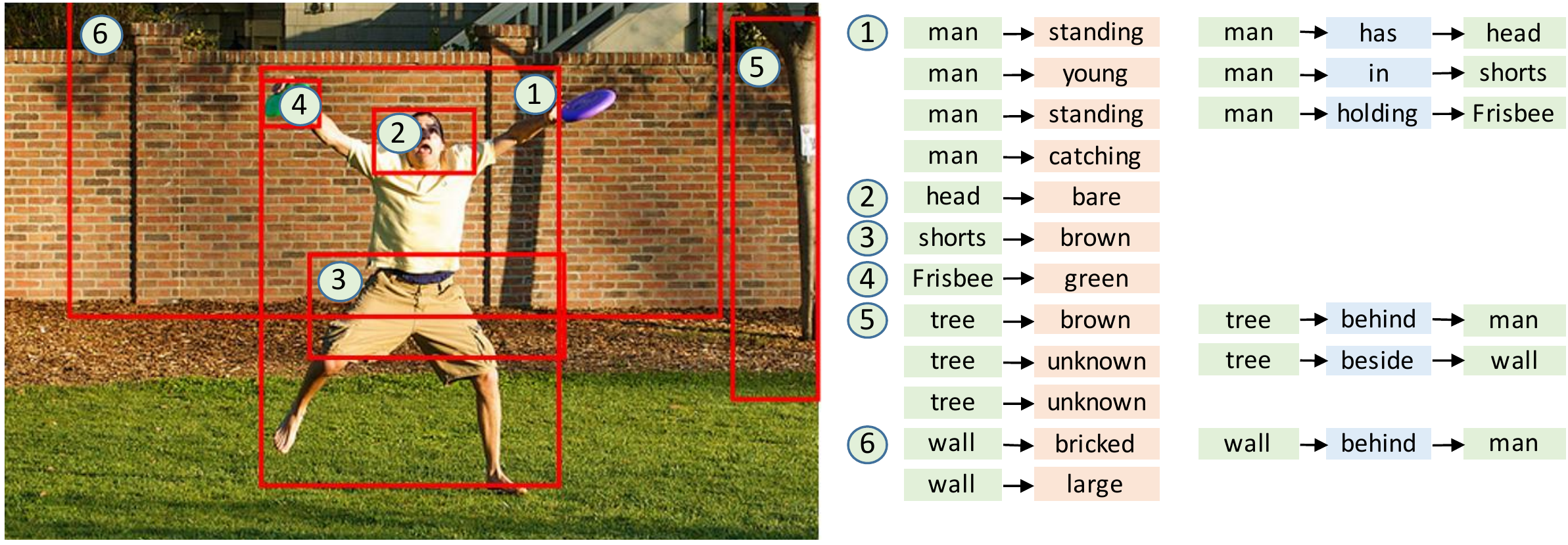}
 		\caption{{{The VRL does a sequential breadth-first search, predicting all relationships and attributes with respect to the current subject instance before moving onto the next instance.} }} 
 		\label{fig:search}
 	\end{center}
 	\vspace{-7mm}
 \end{figure}
 
Instead of learning in the entire action space as in traditional deep RL~\cite{mnih2015human,zhu2016target}, we propose a novel variation-structured traversal scheme over the semantic action graph that dynamically constructs small action sets for each step.

First, VRL uses an object detector to get a set $\mathcal{S}$ of candidate object instances, and then sequentially assigns relationships and attributes to each instance $s \in \mathcal{S}$. For our experiments, we used state-of-the-art Faster R-CNN~\cite{ren2015faster} as the object detector, where the network parameters were initialized using the pre-trained VGG-16 ImageNet model~\cite{simonyan2014very}.

Since subject instances in an image often have multiple relationships and attributes, we do a breadth-first search: we predict all relationships and attributes with respect to the current subject instance of interest, and then move onto the next instance. {We start from the subject instance with the most confident classification score. To prevent the agent from being trapped in a single search path (e.g., in a small local region), the agent selects a new starting subject instance if it has traversed through 5 neighboring objects in the breadth-first search.}

The same object in multiple scenarios may be described by different, semantically ambiguous noun categories that cannot be distinguished by the object detector. To address this semantic ambiguity, we introduce an ambiguity-aware object mining scheme which leverages scene contexts captured by extracted relationships and attributes to help determine the most appropriate object category.

\textbf{Variation-structured action space}. The directed semantic graph $\mathcal{G}$ serves as the action space for VRL. For {any object} instance $s \in \mathcal{S}$ in an image, denote its {object} category by $s_c \in \mathcal{C}$ and its bounding box by $B(s) = (s_{x}, s_{y}, s_w, s_h)$ where $(s_{x}, s_{y})$ is the center coordinate, $s_w$ is the width, and $s_h$ is the height. Given the current subject instance $s$ and { object} instance $s'$, we select three actions $\mathbf{g}_a \in \mathcal{A}$, $\mathbf{g}_p \in \mathcal{P}$, $\mathbf{g}_c \in \mathcal{C}$ according to the VRL network as follows:

(1) Select an attribute $\mathbf{g}_a$ describing $s$ from the set $\Delta_a = \{ a : (s_c, a) \in \mathcal{E}_\mathcal{A} \backslash \mathcal{H}_{\mathcal{A}}(s) \}$, where $\mathcal{H}_{\mathcal{A}}(s)$ denotes the {set of} previously mined attribute {phrases} for $s$.

(2) Select a predicate $\mathbf{g}_p$ relating the subject noun $s_c$ and object noun $s'_c$ {from} $\Delta_p = \{ p : (s_c, p, s'_c) \in \mathcal{E}_\mathcal{P} \}$.

\begin{figure}[!tp]
 	\begin{center}
 		\includegraphics[scale=0.63]{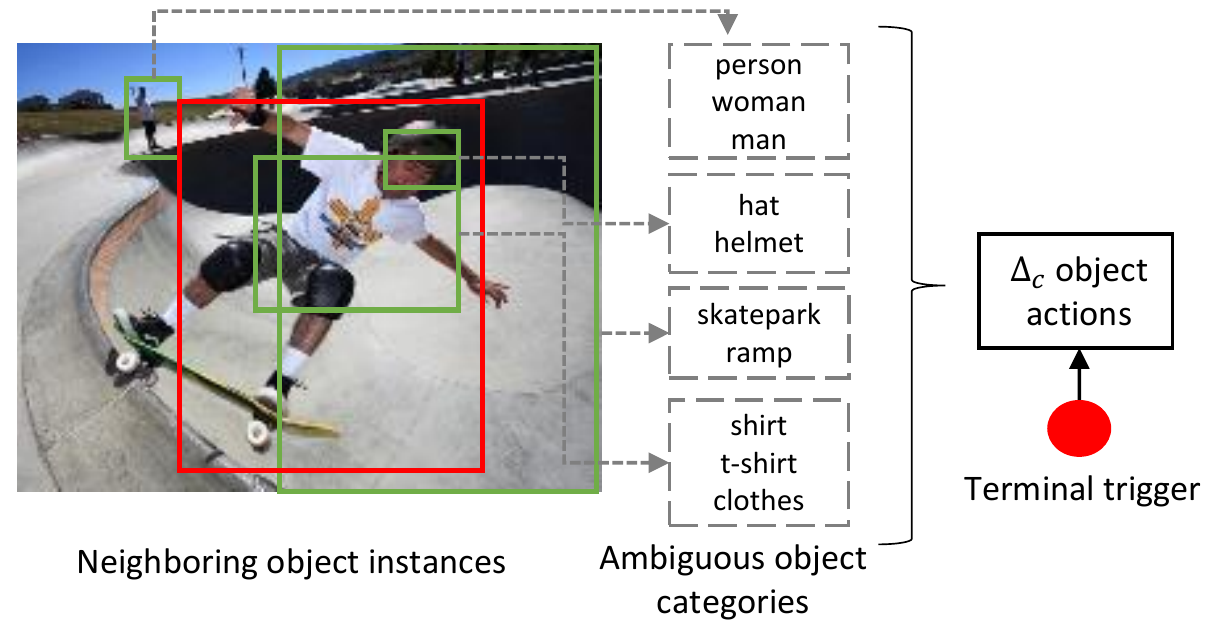}
 		\caption{{Illustration of ambiguity-aware object mining. The image on the left shows the subject instance (red box) and its neighboring object instances (green boxes). The action set $\Delta_c$ contains candidate object categories of each neighboring object which the object detector cannot distinguish (e.g., ``hat'' vs. ``helmet''), and a terminal trigger indicating the end of the object mining scheme for this subject instance. }}
 		\label{fig:objectMine}
 	\end{center}
 	\vspace{-9mm}
 \end{figure}

(3) To select the next object instance $\tilde{s} \in \mathcal{S}$ in the image, we select its corresponding object category $\mathbf{g}_c$ from a set $\Delta_c \subseteq \mathcal{C}$, which is constructed using an ambiguity-aware object mining scheme as follows {(also illustrated in Fig.~\ref{fig:objectMine})}. Let $N(s) \subseteq \mathcal{S}$ be the set of objects neighboring $s$, where a neighbor of $s$ is defined to be any object $\tilde{s} \in \mathcal{S}$ such that $|\tilde{s}_{x}-s_{x}| < 0.5(\tilde{s}_w + s_w)$ and $|\tilde{s}_{y}-s_{y}| < 0.5(\tilde{s}_h + s_h)$. For each object $\tilde{s}$, let $C(\tilde{s})\subseteq \mathcal{C}$ be the set of object categories of $\tilde{s}$ whose confidence scores are at most $0.1$ less than that of the most confident category. Let $\Delta_c = \bigcup_{\tilde{s} \in N(s) \backslash \mathcal{H}_\mathcal{S}} C(\tilde{s}) \cup \{\texttt{\small Terminal} \}$, where $\mathcal{H}_\mathcal{S}$ is the set of previously extracted object instances and \texttt{\small Terminal} is a terminal trigger indicating the end of the object mining scheme for this subject instance. If $N(s) \backslash \mathcal{H}^s$ is empty or the terminal trigger is activated, then we select a new subject instance following the breadth-first scheme. The terminal trigger allows the number of object mining steps for each subject instance to be dynamically specified and limited to a small number.




In each step, the VRL selects actions from the adaptive action sets $\Delta_a, \Delta_p$, and $\Delta_c$, which we call the variation-structured action space due to their dynamic structure.

\textbf{State space.} A detailed overview of the state feature extraction process is shown in Fig.~\ref{fig:network}. Given the current subject $s$ and object $s'$ instances in each time step, the state vector $\mathbf{f}$ is a concatenation of (1) the feature vectors of $s$ and $s'$; (2) the feature vector of the whole image; and (3) a history phrase embedding vector, which is created by concatenating the semantic embeddings of the last two relationship phrases (relating $s$ and $s'$) and the last two attribute phrases (describing $s$) that were mined via the variation-structured traversal scheme. More specifically, each phrase (e.g., ``person riding bicycle'') is embedded into a 2400-dim vector using a pre-trained Skip-thought language model~\cite{kiros2015skip}, thus resulting in a 9600-dim history phrase embedding. 

The feature vector of the whole image provides global context cues which not only help in recognizing relationships and attributes, but also allow the agent to be aware of other uncovered objects. The history phrase embedding captures the search paths and scene contexts that have already been traversed by the agent.

\textbf{Rewards:} Suppose we have groundtruth labels, which consist of the set $\hat{\mathcal{S}}$ of object instances in the image, and attribute phrases $\hat{\mathcal{E}}^{\mathcal{A}}$ and predicate phrases $\hat{\mathcal{E}}^{\mathcal{P}}$ describing the objects in $\hat{\mathcal{S}}$. Given a predicted object instance $s \in \mathcal{S}$, we say that a groundtruth object $\hat{s} \in \hat{\mathcal{S}}$ overlaps with $s$ if they have the same object category (i.e., $s_c = \hat{s}_c \in \mathcal{C}$), and their bounding boxes have at least 0.5 Intersection-over-Union (IoU) overlap.

We define the following reward functions to reflect the detection accuracy of taking action $(\mathbf{g}_a, \mathbf{g}_p, \mathbf{g}_c)$ in state $\mathbf{f}$, where the current subject and object instances are $s$ and $s'$, respectively:

(1) $\mathcal{R}_a(\mathbf{f},\mathbf{g}_a)$ returns +1 if there exists a groundtruth object $\hat{s} \in \hat{\mathcal{S}}$ that overlaps with $s$, and the predicted attribute relationship $(s_c, \mathbf{g}_a)$ is in the groundtruth set $\hat{\mathcal{E}}^{\mathcal{A}}$. Otherwise, it returns -1.

(2) $\mathcal{R}_p(\mathbf{f},\mathbf{g}_p)$ returns +1 if there exists $\hat{s}, \hat{s}' \in \hat{\mathcal{S}}$ that overlap with $s$ and $s'$ respectively, and $(s_c, \mathbf{g}_p, s'_c) \in \hat{\mathcal{E}}^{\mathcal{P}}$. Otherwise, it returns -1.

(3) $\mathcal{R}_c(\mathbf{f},\mathbf{g}_c)$ returns +5 if the next object instance $\tilde{s} \in \mathcal{S}$ corresponding to category $\mathbf{g}_c \in \mathcal{C}$ overlaps with a new groundtruth object $\hat{s} \in \mathcal{S}$. Otherwise, it returns -1. Thus, it encourages faster exploration over all objects in the image.

\subsection{Deep Variation-structured RL}

We optimize three policies to select three actions for each state by maximizing the sum of discounted rewards, which can be formulated as a decision-making process in the deep RL framework. Due to the high-dimensional continuous image data and a model-free environment, we resort to the deep Q-Network (DQN) framework proposed by~\cite{mnih2013playing, mnih2015human}, which generalizes well to unseen inputs. The detailed architecture of our Q-network is illustrated in Fig.~\ref{fig:network}. Specifically, we use DQN to estimate three Q-value sets, parametrized by network weights $\theta_a$, $\theta_p$, $\theta_c$, which correspond to the action sets $\mathcal{A}, \mathcal{P}, \mathcal{C}$.  In each training episode, we use an $\epsilon$-greedy strategy to select actions $\mathbf{g}_a, \mathbf{g}_p, \mathbf{g}_c$ in the variation-structured action space $\Delta_a, \Delta_p, \Delta_c$, where the agent selects random actions with probability $\epsilon$, and selects actions with the highest estimated Q-values with probability $1 - \epsilon$. During testing, we directly select the best actions with highest estimated Q-values in $\Delta_a, \Delta_p, \Delta_c$. The agent sequentially determines the best actions to discover objects, relationships, and attributes in the given image, until either the maximum search step is reached or there are no remaining uncovered object instances.

We also {utilize a replay memory to store experience from past episodes. In each step, we draw a random mini-batch from the replay memory to perform the Q-learning update. The replay memory helps stabilize the training by smoothing the training distribution over past experiences and reducing correlation between training samples~\cite{mnih2013playing, mnih2015human}.} Given {a transition sample} $(\mathbf{f},\mathbf{f}', \mathbf{g}_a, \mathbf{g}_p, \mathbf{g}_c, \mathcal{R}_a, \mathcal{R}_p, \mathcal{R}_c)$, the network weights $\theta^{(t)}_a$, $\theta^{(t)}_p$, $\theta^{(t)}_c$ are updated as follows:
\begin{equation}
\small{
\begin{split}
\theta^{(t+1)}_a = &\theta^{(t)}_a + \alpha(\mathcal{R}_a + \lambda\max_{\mathbf{g}_{a'}}Q(\mathbf{f}', \mathbf{g}_{a'}; \theta^{(t)-}_a) \\
&- Q(\mathbf{f}, \mathbf{g}_{a}; \theta^{(t)}_a))\bigtriangledown_{\theta^{(t)}_a}Q(\mathbf{f}, \mathbf{g}_{a}; \theta^{(t)}_a),\\
\theta^{(t+1)}_p = &\theta^{(t)}_p + \alpha(\mathcal{R}_p + \lambda\max_{\mathbf{g}_{p'}}Q(\mathbf{f}', \mathbf{g}_{p'}; \theta^{(t)-}_p) \\
&- Q(\mathbf{f}, \mathbf{g}_{p}; \theta^{(t)}_p))\bigtriangledown_{\theta^{(t)}_p}Q(\mathbf{f}, \mathbf{g}_{p}; \theta^{(t)}_p),\\
\theta^{(t+1)}_c = &\theta^{(t)}_c + \alpha(\mathcal{R}_c + \lambda\max_{\mathbf{g}_{c'}}Q(\mathbf{f}', \mathbf{g}_{c'}; \theta^{(t)-}_c) \\
&- Q(\mathbf{f}, \mathbf{g}_{c}; \theta^{(t)}_c))\bigtriangledown_{\theta^{(t)}_c}Q(\mathbf{f}, \mathbf{g}_{c}; \theta^{(t)}_c),
\end{split}}
\end{equation}
where $\mathbf{g}_{a'}$, $\mathbf{g}_{p'}$, $\mathbf{g}_{c'}$ represent the actions that can be taken in state $\mathbf{f}'$, $\alpha$ is the learning {rate,} and $\lambda$ is the discount factor. The target network weights {$\theta^{(t)-}_a$, $\theta^{(t)-}_p$, $\theta^{(t)-}_c$ are copied every $\tau$ steps from the online network, and kept fixed in all other steps.}

\section{Experiments}
\label{sec:exp}
\textbf{Dataset.} We conduct our experiments on the Visual Relationship Detection (VRD) dataset~\cite{lu2016visual} and the Visual Genome dataset~\cite{krishna2016visual}. VRD~\cite{lu2016visual} contains 5000 images (4000 for training, 1000 for testing) with 100 object categories and 70 predicates. In total, the dataset contains 37,993 relationship instances with 6,672 relationship types, out of which 1,877 relationships occur only in the test set and not in the training set. For the Visual Genome Dataset~\cite{krishna2016visual}, we experiment on 87,398 images (out of which 5000 are held out for validation, and 5000 for testing), containing 703,839 relationship instances with 13,823 relationship types and 1,464,446 attribute instances with 8,561 attribute types. There are 2,015 relationship types that occur in the test set but not in the training set, which allows us to evaluate VRL on zero-shot learning.

\textbf{Implementation Details.} We train a deep Q-network for 60 epochs with a shared RMSProp optimizer~\cite{tieleman2012lecture}. Each epoch ends after performing an episode on all training {images}. We use a mini-batch size of 64 images. The { maximum} search step for each {image} is empirically set {to} 300. During $\epsilon$-greedy training, $\epsilon$ is annealed linearly from 1 to 0.1 over the first 20 epochs, and is fixed to 0.1 in the remaining epochs. The discount factor $\lambda$ is set to 0.9, and the network parameters $\theta^{(t)-}_a$, $\theta^{(t)-}_p$ and $\theta^{(t)-}_c$ are copied after every $\tau=10000$ steps. The learning rate $\alpha$ is initialized to 0.0007 and decreased by a factor of 10 after every 10 epochs. {Only} the top 100 candidate object instances, {ranked by objectness confidence scores by the trained object detector}, are selected for mining relationships and attributes in an image, in order to {balance} efficiency and effectiveness. On VRD~\cite{lu2016visual}, VRL takes about 8 hours to train an object detector with 100 object categories, and two days to converge. On the Visual Genome dataset~\cite{krishna2016visual}, VRL takes between 4 to 5 days to train an object detector with 1750 object categories, and one week to converge. On average, it takes 300ms to feed-forward one image into VRL. More details about the dataset are provided in Sec.~\ref{sec:exp}. The implementations are based on the publicly available Torch7 platform on a single NVIDIA GeForce GTX 1080.

\begin{table}[!tp]\setlength{\tabcolsep}{2pt}
	\centering\scriptsize
	\caption{Results for relationship phrase detection (Phr.) and relationship detection (Rel.) on the VRD dataset. R@100 and R@50 are abbreviations for Recall@100 and Recall@50.}\label{tab:VRD}
	\vspace{1pt}
	\begin{tabular}{c|c|c|c|cccccccccccccccccc}
		\toprule
		{Method} &  Phr. R@100 & Phr. R@50  &  Rel. R@100 & Rel. R@50  \\
		\midrule			
		Visual Phrases~\cite{sadeghi2011recognition} & 0.07 & 0.04 & - & - \\
		\midrule
		{Joint CNN+R-CNN~\cite{simonyan2014very}} & {0.09} & 0.07 & 0.09 & 0.07  \\
		\midrule
		Joint CNN+RPN~\cite{simonyan2014very} & {2.18} & 2.13 & 1.17 & 1.15  \\
		\midrule
		Lu et al. V only ~\cite{lu2016visual} & {2.61} & 2.24 & 1.85 & 1.58  \\
		\midrule
		Faster R-CNN~\cite{ren2015faster} & {3.31} & 3.24 & - & -  \\	
		\midrule
		Joint CNN+Trained RPN~\cite{ren2015faster} & 3.51 & 3.17 & 2.22 & 1.98  \\
		
		\midrule
		Faster R-CNN V only ~\cite{ren2015faster} & {6.13} & 5.61 & 5.90 & 4.26  \\
		
		\midrule
		Lu et al.~\cite{lu2016visual} & {17.03} & 16.17 & 14.70 & 13.86  \\
		
		\midrule
		\textbf{Our VRL} & \textbf{22.60} & \textbf{21.37} & \textbf{20.79} & \textbf{18.19}  \\
		\hline
		\midrule
		Lu et al.~\cite{lu2016visual} (zero-shot) & {3.76} & 3.36 & 3.28 & 3.13  \\
		\midrule
		{Our VRL} (zero-shot) & {10.31} & {9.17} & {8.52} & {7.94}  \\
		\bottomrule
	\end{tabular}%
	\vspace{-6mm}
\end{table}%

\textbf{Evaluation.} Following~\cite{lu2016visual}, we use recall@100 and recall@50 as our evaluation metrics. Recall@$x$ computes the fraction of times the correct relationship or attribute instance is covered in the top $x$ confident predictions, which are ranked by the {product of objectness confidence scores for the relevant} object instances (i.e., confidence scores of the object detector) and Q-values of the selected predicates or attributes. As discussed in~\cite{lu2016visual}, we do not use the mean average precision (mAP), which is a pessimistic evaluation metric because the dataset cannot exhaustively annotate all possible relationships and attributes in an image.

Following~\cite{lu2016visual}, we evaluate on three tasks: (1) In relationship phrase detection~\cite{sadeghi2011recognition}, the goal is to predict a ``subject-predicate-object'' phrase, where the localization of the entire relationship has at least 0.5 overlap with a groundtruth bounding box. (2) In relationship detection, the goal is to predict a ``subject-predicate-object'' phrase, where the localizations of the subject and object instances have at least 0.5 overlap with their corresponding groundtruth boxes. (3) In attribute detection, the goal is to predict a ``subject-attribute'' phrase, where the subject's localization has at least 0.5 overlap with a groundtruth box.

\textbf{Baseline models.} First, we compare our model with {state-of-the-art} approaches, \textbf{Visual Phrases~\cite{sadeghi2011recognition}}, \textbf{Joint CNN+R-CNN~\cite{simonyan2014very}} and \textbf{Lu et al.~\cite{lu2016visual}}. {Note that the latter two methods use R-CNN~\cite{girshick2014rich} to extract object proposals.} Their results on VRD are reported in~\cite{lu2016visual}, and we also experiment their methods on the Visual Genome dataset. \textbf{Lu et al. V only ~\cite{lu2016visual}} {trains} individual detectors for object and predicate categories separately, and then {combines} their confidences to generate {a} relationship prediction. Furthermore, we train and compare with the following models: ``\textbf{Faster R-CNN~\cite{ren2015faster}}'' directly detects each unique relationship or attribute type, following Visual Phrases~\cite{sadeghi2011recognition}. ``\textbf{Faster R-CNN V only~\cite{ren2015faster}}'' model is similar to {Lu et al. V only ~\cite{lu2016visual}}, with the only difference being that Faster R-CNN is used for object detection. ``\textbf{Joint CNN+RPN~\cite{simonyan2014very}}'' extracts proposals using the pre-trained RPN~\cite{ren2015faster} model on VOC 2012~\cite{everingham2010pascal} and then performs the classification. ``\textbf{Joint CNN+Trained RPN~\cite{ren2015faster}}'' trains a separate RPN model on our dataset to generate proposals.

\subsection{Comparison with State-of-the-art Models} 

\begin{table}[!tp]\setlength{\tabcolsep}{2pt}
	\centering\scriptsize
	\caption{Results for relationship detection on Visual Genome. }\label{tab:VG_rel}
	\begin{tabular}{c|c|c|c|cccccccccccccccccc}
		\toprule
		{Method} &   Phr. R@100 & Phr. R@50  &  Rel. R@100 & Rel. R@50  \\

		\midrule
		{Joint CNN+R-CNN~\cite{simonyan2014very}} & {0.13} & 0.10 & 0.11 & 0.08  \\
		\midrule
		Joint CNN+RPN~\cite{simonyan2014very} & {1.39} & 1.34 & 1.22 & 1.18  \\	
		\midrule
		Lu et al. V only~\cite{lu2016visual} & {1.66} & {1.54} & {1.48} & {1.20}  \\
			
		\midrule
		Faster R-CNN~\cite{ren2015faster} & {2.25} & 2.19 & - & -  \\		
		\midrule
		Joint CNN+Trained RPN~\cite{ren2015faster} & {2.52} & 2.44 & 2.37 & 2.23  \\
	
		\midrule
		Faster R-CNN V only~\cite{ren2015faster} & {5.79} & {5.22} & {4.87} & {4.36}  \\
		\midrule
		Lu et al.~\cite{lu2016visual} & {10.23} & {9.55} & {7.96} & {6.01}  \\
		
		\midrule
		\textbf{Our VRL} & \textbf{16.09} & \textbf{14.36} & \textbf{13.34} & \textbf{12.57}  \\
		\hline
		\midrule
		Lu et al.~\cite{lu2016visual} (zero-shot) & {1.20} & 1.08 & 1.13 & 0.97  \\
		\midrule
		{Our VRL} (zero-shot) & {7.98} & {6.53} & {7.14} & {6.27}  \\
		\bottomrule
	\end{tabular}%
	\vspace{-6mm}
\end{table}%

\begin{table}[!tp]\setlength{\tabcolsep}{2pt}
	\centering\scriptsize
	\caption{Results for attribute detection on Visual Genome. }\label{tab:vg_attr}
	\begin{tabular}{c|c|cccccccccccccccccccc}
		\toprule
		{Method} &  Attribute Recall@100 & Attribute Recall@50  \\
		\midrule
		{Joint CNN+R-CNN~\cite{simonyan2014very}} & {2.38} & 1.97   \\
		\midrule
		Joint CNN+RPN~\cite{simonyan2014very} & {3.48} & 2.63 \\
		\midrule
		Faster R-CNN~\cite{ren2015faster} & {7.36} & 5.22 \\
		\midrule
		Joint CNN+Trained RPN~\cite{ren2015faster} & {9.77} & 8.35 \\
		
		\midrule
		\textbf{Our VRL} & \textbf{26.43} & \textbf{24.87} \\
		\bottomrule
	\end{tabular}%
	\vspace{-6mm}
\end{table}%

Comparison of results with baseline methods on VRD and {Visual} Genome are reported in Tables~\ref{tab:VRD}, \ref{tab:VG_rel}, and \ref{tab:vg_attr}.

\textbf{Shared Detectors vs Individual Detectors.} The compared models can be categorized into two classes: (1) {Models that train individual detectors for each predicate or attribute type, i.e.,} Visual Phrases~\cite{sadeghi2011recognition}, Joint CNN+R-CNN~\cite{simonyan2014very}, Joint CNN+RPN~\cite{simonyan2014very}, Faster R-CNN~\cite{ren2015faster}, Joint CNN+Trained RPN~\cite{ren2015faster}. (2) Models that train shared detectors for predicate or attribute types, and then combine their results with object detectors to generate the final prediction, i.e., Lu et al. V only~\cite{lu2016visual}, Faster R-CNN V only~\cite{ren2015faster}, Lu et al.~\cite{lu2016visual} and our VRL. Since the space of all possible relationships and attributes is often large, there are insufficient training examples for infrequent relationships, leading to poor average performance of the models that use individual detectors.

\textbf{RPN vs R-CNN.} In all cases, we obtain performance improvements using RPN network~\cite{ren2015faster} {over R-CNN}~\cite{girshick2014rich} for proposal generation. Additionally, training the proposal network on VRD and VG datasets can also increase the recalls over the pre-trained networks on other datasets.

\begin{figure}[!tp]
	\begin{center}
		\includegraphics[scale=0.38]{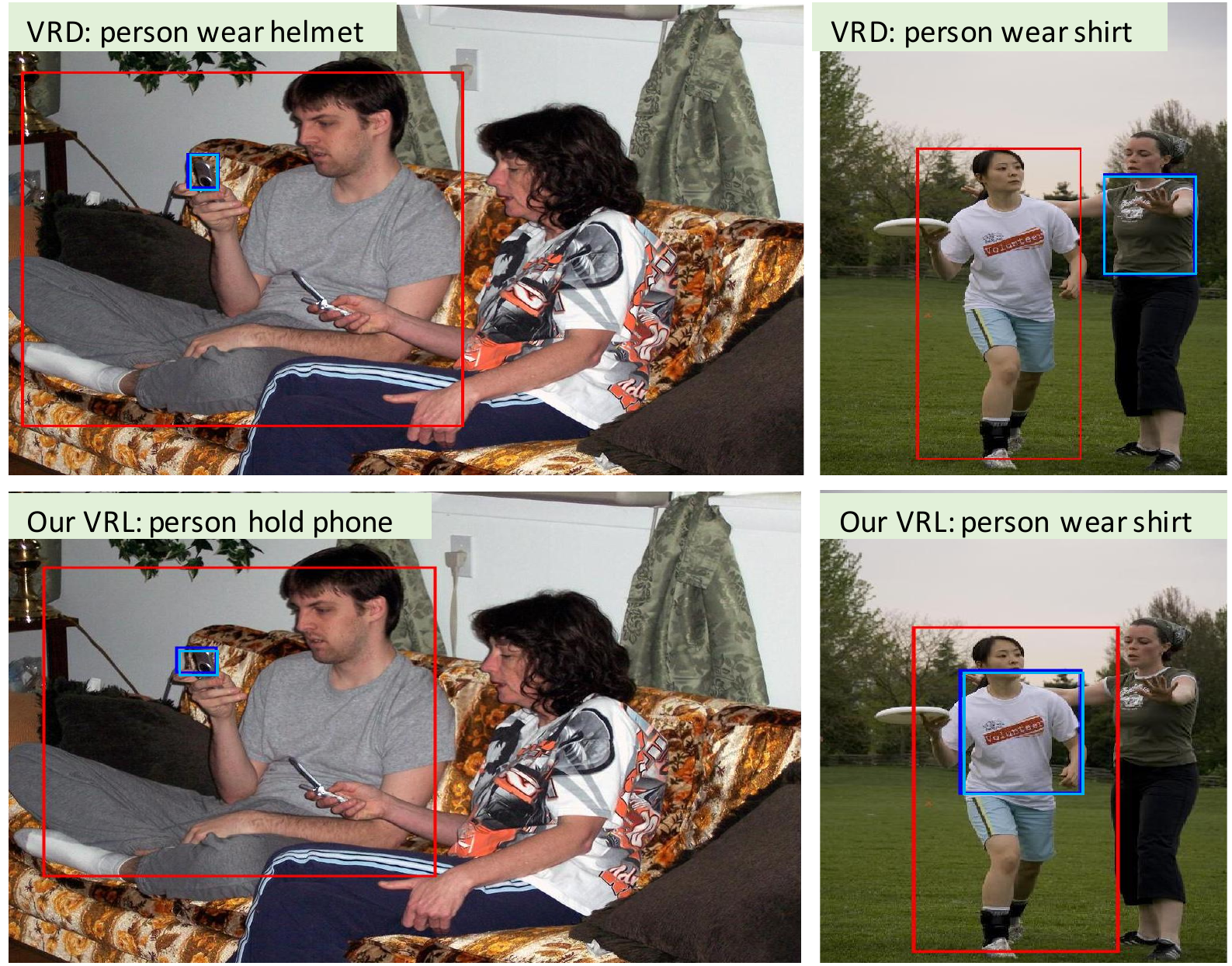}
		\caption{{Qualitative comparison between VRD~\cite{lu2016visual} and our VRL.}}
		\label{fig:comparison}
	\end{center}
	\vspace{-8mm}
\end{figure}

\textbf{Language Priors.} Unlike baselines that simply train classifiers from visual cues, VRL and Lu et al.~\cite{lu2016visual} leverage language priors to facilitate prediction. Lu et al.~\cite{lu2016visual} uses semantic word embeddings to finetune the likelihood of a predicted relationship, while VRL {follows} a variational-structured traversal scheme over a directed semantic action graph {built from language priors}. Both VRL and Lu et al.~\cite{lu2016visual} achieve significantly better performance than other baselines, which demonstrates the necessity of language priors for relationship and attribute detection. Moreover, VRL still shows substantial improvements comparison to Lu et al.~\cite{lu2016visual}. Therefore, VRL's directed semantic action graph provides a more compact and rich representation of semantic correlations than the word embeddings used in Lu et al.~\cite{lu2016visual}. The significant performance improvement is also {due to the sequential reasoning of RL.}

\begin{figure*}[!tp]
	\begin{center}
		\includegraphics[scale=0.47]{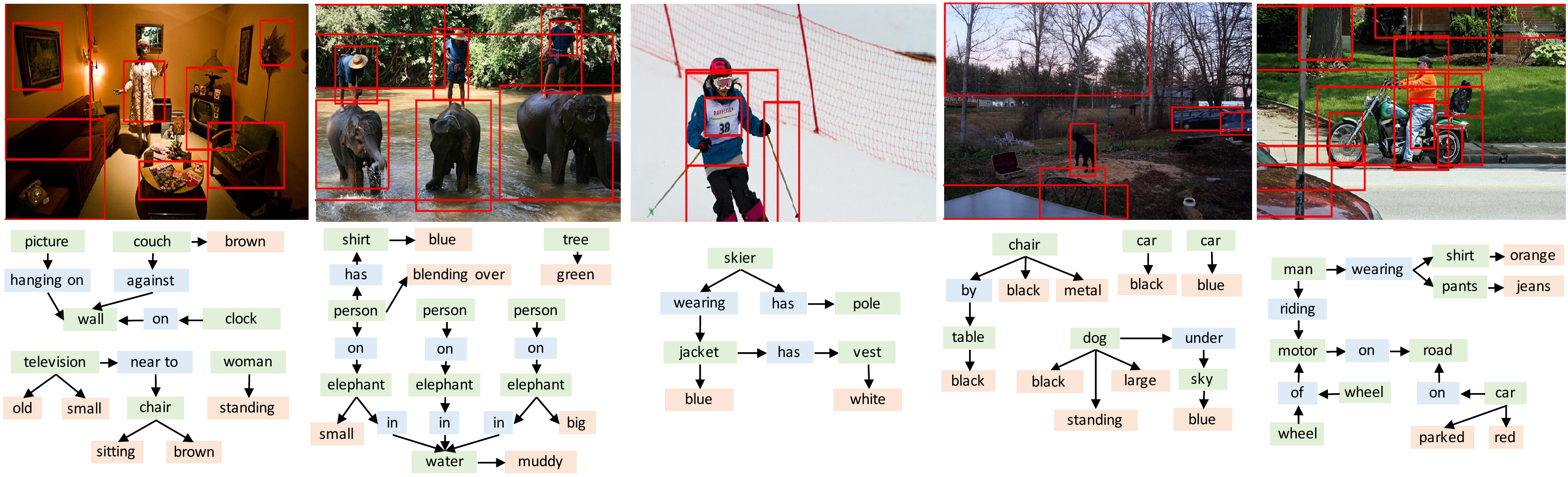}
		\caption{{Examples of relationship and attribute detection results generated by {VRL} on the Visual Genome dataset. We show the top predictions for each image: the localized objects (top) and a semantic graph describing their relationships and attributes (bottom).}}
		\label{fig:results}
	\end{center}
	\vspace{-8mm}
\end{figure*}

\textbf{Qualitative Comparisons} We show some qualitative comparison with Lu et al.~\cite{lu2016visual} in Fig.~\ref{fig:comparison}, and more detection results of VRL in Fig.~\ref{fig:results}. Our VRL generates a rich understanding of the image, including the localization and recognition of objects, and the detection of object relationships and attributes. For instance, VRL can correctly detect interactions (``person on elephant'', ``man riding motor''), spatial layouts (``picture hanging on wall'', ``car on road''), parts of objects (``person has shirt'', ``wheel of motor''), and attribute descriptions (``television old'', ``woman standing''). 

\subsection{Discussion}
We give further analysis on the key components of VRL, and report the results in Table~\ref{tab:variants}.

\textbf{Reinforcement Learning vs. Random Walk.} The variant ``\textbf{RL}'' is a standard deep RL model that selects three actions over the entire action space instead of the variation-structured action space.  We compare RL with a simple ``\textbf{Random Walk}'' traversal scheme where in each step, the agent randomly selects one object instance in the image, and predicts relationships and attributions for the two most-recently selected instances.``Random Walk'' only achieves slightly better results than ``Joint+Trained RPN~\cite{ren2015faster}'' and performs much worse than the remaining variants, again demonstrating the benefit of sequential mining in RL. 

\textbf{Variation-structured traversal scheme.} VRL achieves a remarkably higher recall compared to RL (e.g., 13.34\% vs 6.23\% on relationship detection and 26.43\% vs 12.47\% on attribute detection, in terms of Recall@100). Thus, we conclude that using a variation-structured traversal scheme to dynamically configure the small action set for each state can accelerate and stabilize the learning procedure, by dramatically {decreasing} the number of possible actions. For example, the number of predicate actions (347) can be dropped to 15 on average.
\begin{figure}[!tp]
	\begin{center}
		\includegraphics[scale=0.65]{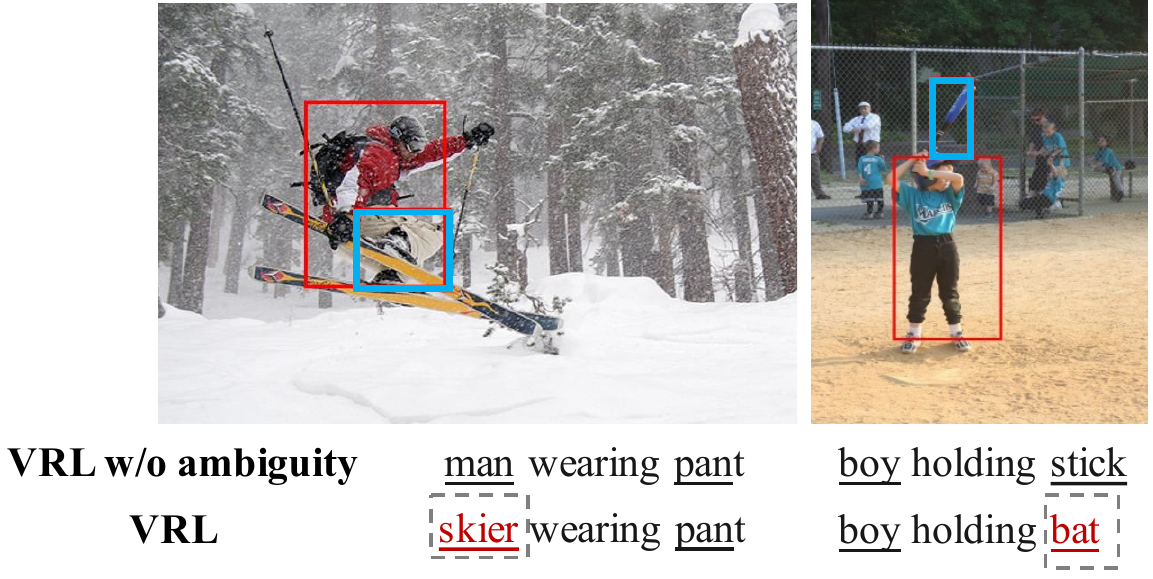}
		\caption{{Comparison between ``VRL w/o ambiguity'' and VRL. Using ambiguity-aware object mining, VRL successfully resolves vague predictions into more concrete ones {(``man'' $\rightarrow$ ``skier'' and ``stick'' $\rightarrow$ ``bat'')}.}}
		\label{fig:ambiguity}
	\end{center}
	\vspace{-8mm}
\end{figure}

\begin{table}[!tp]\setlength{\tabcolsep}{2pt}
	\centering\scriptsize
	\caption{Performance {of VRL} and its variants on Visual Genome.}\label{tab:variants}
	\begin{tabular}{c|c|c|c|cccccccccccccccccc}
		\toprule
		{Method} &  Rel. R@100 & Rel. R@50 & Attr. R@100 & Attr. R@50 \\
		\midrule
		Joint CNN+Trained RPN~\cite{ren2015faster} & {2.37} & 2.23 & {9.77} & 8.35\\
		\midrule
		Random Walk & {3.67} & {3.09} & {10.21} & {8.59}\\
		\midrule
		RL & {6.23} & {5.10} & {12.47} & {10.09}\\
		\midrule
		{VRL w/o history phrase} & {9.05} & {8.12} & {20.09} & {19.45}  \\
		\midrule
		{VRL w/ directional actions} & {10.66} & {9.85} & {20.31} & {18.62} 
		\\ 
		\midrule
		{VRL {w/ historical} actions} & {11.98} & {10.01} & {23.02} & {22.15}  \\
		\midrule
		{VRL w/o ambiguity} & {12.01} & {11.20} & {24.78} & {22.46}  \\
		\midrule
		\textbf{Our VRL} & \textbf{13.34} & \textbf{12.57} & \textbf{26.43} & \textbf{24.87}  \\
		\midrule
		{{VRL} w/ LSTM} & {13.86} & {13.07} & {25.98} & {25.01}  \\
		\bottomrule
	\end{tabular}%
	\vspace{-6mm}
\end{table}%

\textbf{History phrase embedding.}  To validate the effectiveness of history phrase {embeddings}, we evaluate two variants of VRL: (1) ``\textbf{VRL w/o history phrase}'' does not incorporate history phrase embedding into state features. This variant causes the recall to drop by over 4\% compared to the original VRL. Thus, leveraging history phrase embeddings can help inform the current state what has happened in the past and stabilize search trajectories that might get stuck in repetitive cycles. (2) ``\textbf{VRL w/ {historical} actions}'' directly stores a historical action vector in the state~\cite{caicedo2015active}. Each historical action vector is the concatenation of four ($|\mathcal{C}|+|\mathcal{A}|+|\mathcal{P}|$)-dim action vectors corresponding to the last four actions taken, where each action vector is zero in all elements except the indices corresponding to the three actions taken in $\mathcal{C}$, $\mathcal{A}$, $\mathcal{P}$. This variant still causes the recall to drop, demonstrating that semantic phrase embeddings learned by language models can capture richer history cues (e.g., relationship similarity).

\textbf{Ambiguity-aware object mining.} ``\textbf{VRL w/o ambiguity}'' only considers the top-1 predicted category of each object for the action set $\Delta_c$. It obtains lower recall than VRL, suggesting that incorporating {semantically} ambiguous categories into $\Delta_c$ can help identify a more appropriate category for each object under different scene contexts. Fig.~\ref{fig:ambiguity} illustrates two examples where VRL successfully resolves vague predictions of ``VRL w/o ambiguity'' into more concrete ones (``man''$\rightarrow$``skier'' and ``stick''$\rightarrow$``bat'').
 
\textbf{Spatial actions.} Similar to \cite{caicedo2015active}, we experiment using spatial actions in the deep RL setting to sequentially extract object instances.  The variant ``\textbf{VRL w/ directional actions}'' replaces the 1751-dim object category action vector with a 9-dim action vector indexed by directions (N, NE, E, SE, S, SW, W, NW) plus one terminal trigger. In each step, the agent selects a neighboring object instance with the highest confidence whose center lies in one of the eight directions w.r.t. that of the subject instance. The diverse spatial layouts of object instances across different images make it difficult to learn a spatial action policy, and causes this variant to perform poorly.


{\textbf{Long Short-Term Memory} ``\textbf{VRL w/ LSTM}'' is a variant where all fully-connected layers in Fig.~\ref{fig:network} are replaced with LSTM~\cite{hochreiter1997long} layers, which have shown promising results in capturing long-term dependencies.} However, ``VRL w/ LSTM'' has no noticeable performance improvements over VRL, while requiring much more training time. {This shows that history phrase embeddings can sufficiently} model {historical} context for sequential prediction.

\subsection{Zero-shot Learning} We also compare VRL with Lu et al.~\cite{lu2016visual} in the zero-shot learning setting (see Tables~\ref{tab:VRD} and \ref{tab:VG_rel}). A promising model should be capable of predicting unseen relationships, since the training data will not cover all possible relationship types. Lu et al.~\cite{lu2016visual} uses word embeddings to project similar relationships onto unseen ones, while our VRL uses a large semantic action graph to learn similar relationships on shared graph nodes. Our VRL achieves $>5\%$ performance improvements over Lu et al.~\cite{lu2016visual} on both datasets.

\section{Conclusion and Future Work}

We proposed a novel deep variation-structured reinforcement learning framework for detecting visual relationships and attributes. The VRL sequentially discovers the relationship and attribute instances following a variation-structured traversal scheme on a directed semantic action graph. It incorporates global interdependency to facilitate predictions in local regions. Our experiments on VRD and the Visual Genome dataset demonstrate the power and efficiency of our model over baselines. As future work, a larger directed action graph can be built using natural language sentences. Additionally, VRL can be generalized into an unsupervised learning framework to learn from a massive number of unlabeled images.

{\small
\bibliographystyle{ieee}
\bibliography{egbib}

\begin{thebibliography}{10}\itemsep=-1pt

\bibitem{antol2015vqa}
S.~Antol, A.~Agrawal, J.~Lu, M.~Mitchell, D.~Batra, C.~Lawrence~Zitnick, and
  D.~Parikh.
\newblock Vqa: Visual question answering.
\newblock In {\em Proceedings of the IEEE International Conference on Computer
  Vision}, pages 2425--2433, 2015.

\bibitem{caicedo2015active}
J.~C. Caicedo and S.~Lazebnik.
\newblock Active object localization with deep reinforcement learning.
\newblock In {\em Proceedings of the IEEE International Conference on Computer
  Vision}, pages 2488--2496, 2015.

\bibitem{everingham2010pascal}
M.~Everingham, L.~Van~Gool, C.~K. Williams, J.~Winn, and A.~Zisserman.
\newblock The pascal visual object classes (voc) challenge.
\newblock {\em International journal of computer vision}, 88(2):303--338, 2010.

\bibitem{farhadi2009describing}
A.~Farhadi, I.~Endres, D.~Hoiem, and D.~Forsyth.
\newblock Describing objects by their attributes.
\newblock In {\em IEEE Conference on Computer Vision and Pattern Recognition
  (CVPR)}, pages 1778--1785, 2009.

\bibitem{girshick2014rich}
R.~Girshick, J.~Donahue, T.~Darrell, and J.~Malik.
\newblock Rich feature hierarchies for accurate object detection and semantic
  segmentation.
\newblock In {\em Proceedings of the IEEE conference on computer vision and
  pattern recognition}, pages 580--587, 2014.

\bibitem{gu2016deep}
S.~Gu, E.~Holly, T.~Lillicrap, and S.~Levine.
\newblock Deep reinforcement learning for robotic manipulation.
\newblock {\em arXiv preprint arXiv:1610.00633}, 2016.

\bibitem{he2015deep}
K.~He, X.~Zhang, S.~Ren, and J.~Sun.
\newblock Deep residual learning for image recognition.
\newblock In {\em The IEEE Conference on Computer Vision and Pattern
  Recognition (CVPR)}, June 2016.

\bibitem{hochreiter1997long}
S.~Hochreiter and J.~Schmidhuber.
\newblock Long short-term memory.
\newblock {\em Neural computation}, 9(8):1735--1780, 1997.

\bibitem{johnson2015densecap}
J.~Johnson, A.~Karpathy, and L.~Fei-Fei.
\newblock Densecap: Fully convolutional localization networks for dense
  captioning.
\newblock In {\em IEEE Conference on Computer Vision and Pattern Recognition
  (CVPR)}, June 2016.

\bibitem{johnson2015image}
J.~Johnson, R.~Krishna, M.~Stark, L.-J. Li, D.~A. Shamma, M.~S. Bernstein, and
  L.~Fei-Fei.
\newblock Image retrieval using scene graphs.
\newblock In {\em IEEE Conference on Computer Vision and Pattern Recognition
  (CVPR)}, pages 3668--3678, 2015.

\bibitem{kaelbling1996reinforcement}
L.~P. Kaelbling, M.~L. Littman, and A.~W. Moore.
\newblock Reinforcement learning: A survey.
\newblock {\em Journal of artificial intelligence research}, 4:237--285, 1996.

\bibitem{kiros2015skip}
R.~Kiros, Y.~Zhu, R.~R. Salakhutdinov, R.~Zemel, R.~Urtasun, A.~Torralba, and
  S.~Fidler.
\newblock Skip-thought vectors.
\newblock In {\em Advances in neural information processing systems}, pages
  3294--3302, 2015.

\bibitem{krishna2016visual}
R.~Krishna, Y.~Zhu, O.~Groth, J.~Johnson, K.~Hata, J.~Kravitz, S.~Chen,
  Y.~Kalantidis, L.-J. Li, D.~A. Shamma, et~al.
\newblock Visual genome: Connecting language and vision using crowdsourced
  dense image annotations.
\newblock {\em arXiv preprint arXiv:1602.07332}, 2016.

\bibitem{liao2016support}
W.~Liao, M.~Y. Yang, H.~Ackermann, and B.~Rosenhahn.
\newblock On support relations and semantic scene graphs.
\newblock {\em arXiv preprint arXiv:1609.05834}, 2016.

\bibitem{long2015fully}
J.~Long, E.~Shelhamer, and T.~Darrell.
\newblock Fully convolutional networks for semantic segmentation.
\newblock In {\em Proceedings of the IEEE Conference on Computer Vision and
  Pattern Recognition}, pages 3431--3440, 2015.

\bibitem{lu2016visual}
C.~Lu, R.~Krishna, M.~Bernstein, and L.~Fei-Fei.
\newblock Visual relationship detection with language priors.
\newblock In {\em European Conference on Computer Vision}, pages 852--869.
  Springer, 2016.

\bibitem{mnih2013playing}
V.~Mnih, K.~Kavukcuoglu, D.~Silver, A.~Graves, I.~Antonoglou, D.~Wierstra, and
  M.~Riedmiller.
\newblock Playing atari with deep reinforcement learning.
\newblock In {\em Deep Learning, Neural Information Processing Systems
  Workshop}, 2013.

\bibitem{mnih2015human}
V.~Mnih, K.~Kavukcuoglu, D.~Silver, A.~A. Rusu, J.~Veness, M.~G. Bellemare,
  A.~Graves, M.~Riedmiller, A.~K. Fidjeland, G.~Ostrovski, et~al.
\newblock Human-level control through deep reinforcement learning.
\newblock {\em Nature}, 518(7540):529--533, 2015.

\bibitem{qin2013query}
D.~Qin, C.~Wengert, and L.~Van~Gool.
\newblock Query adaptive similarity for large scale object retrieval.
\newblock In {\em Proceedings of the IEEE Conference on Computer Vision and
  Pattern Recognition}, pages 1610--1617, 2013.

\bibitem{ren2015faster}
S.~Ren, K.~He, R.~Girshick, and J.~Sun.
\newblock Faster r-cnn: Towards real-time object detection with region proposal
  networks.
\newblock In {\em Advances in neural information processing systems}, pages
  91--99, 2015.

\bibitem{sadeghi2015viske}
F.~Sadeghi, S.~K. Divvala, and A.~Farhadi.
\newblock Viske: Visual knowledge extraction and question answering by visual
  verification of relation phrases.
\newblock In {\em IEEE Conference on Computer Vision and Pattern Recognition
  (CVPR)}, pages 1456--1464, 2015.

\bibitem{sadeghi2011recognition}
M.~A. Sadeghi and A.~Farhadi.
\newblock Recognition using visual phrases.
\newblock In {\em IEEE Conference on Computer Vision and Pattern Recognition
  (CVPR)}, pages 1745--1752, 2011.

\bibitem{schuster2015generating}
S.~Schuster, R.~Krishna, A.~Chang, L.~Fei-Fei, and C.~D. Manning.
\newblock Generating semantically precise scene graphs from textual
  descriptions for improved image retrieval.
\newblock In {\em Proceedings of the Fourth Workshop on Vision and Language},
  pages 70--80. Citeseer, 2015.

\bibitem{silver2016mastering}
D.~Silver, A.~Huang, C.~J. Maddison, A.~Guez, L.~Sifre, G.~Van Den~Driessche,
  J.~Schrittwieser, I.~Antonoglou, V.~Panneershelvam, M.~Lanctot, et~al.
\newblock Mastering the game of go with deep neural networks and tree search.
\newblock {\em Nature}, 529(7587):484--489, 2016.

\bibitem{simonyan2014very}
K.~Simonyan and A.~Zisserman.
\newblock Very deep convolutional networks for large-scale image recognition.
\newblock In {\em ICLR}, 2015.

\bibitem{tieleman2012lecture}
T.~Tieleman and G.~Hinton.
\newblock Lecture 6.5-rmsprop: Divide the gradient by a running average of its
  recent magnitude.
\newblock {\em COURSERA: Neural Networks for Machine Learning}, 4(2), 2012.

\bibitem{zhu2014reasoning}
Y.~Zhu, A.~Fathi, and L.~Fei-Fei.
\newblock Reasoning about object affordances in a knowledge base
  representation.
\newblock In {\em European Conference on Computer Vision}, pages 408--424.
  Springer, 2014.

\bibitem{zhu2016target}
Y.~Zhu, R.~Mottaghi, E.~Kolve, J.~J. Lim, A.~Gupta, L.~Fei-Fei, and A.~Farhadi.
\newblock Target-driven visual navigation in indoor scenes using deep
  reinforcement learning.
\newblock {\em arXiv preprint arXiv:1609.05143}, 2016.

\end{thebibliography}
}

\end{document}